\documentclass{article}
\usepackage{spconf,amsmath,graphicx,hyperref}

\title{AMPose: Alternately Mixed Global-Local Attention Model for 3D Human Pose Estimation}
\name{Hongxin Lin, Yunwei Chiu and Peiyuan Wu}
\address{National Taiwan University, Taiwan
}
\begin{document}
%
\maketitle
\begin{abstract}
The graph convolutional networks (GCNs) have been applied to model the physically connected and non-local relations among human joints for 3D human pose estimation (HPE). In addition, the purely Transformer-based models recently show promising results in video-based 3D HPE. However, the single-frame method still needs to model the physically connected relations among joints because the feature representations transformed only by global relations via the Transformer neglect information on the human skeleton. To deal with this problem, we propose a novel method in which the Transformer encoder and GCN blocks are alternately stacked, namely AMPose, to combine the global and physically connected relations among joints towards HPE. In the AMPose, the Transformer encoder is applied to connect each joint with all the other joints, while GCNs are applied to capture information on physically connected relations. The effectiveness of our proposed method is evaluated on the Human3.6M dataset. Our model also shows better generalization ability by testing on the MPI-INF-3DHP dataset. Code can be retrieved at \url{https://github.com/erikervalid/AMPose}.
\vspace{-0.1cm}
\end{abstract}

\section{Introduction}
\label{sec:intro}
Human pose estimation (HPE) is attractive to researchers in computer vision. In particular, 3D HPE is rather related to real-world applications in human-robot interaction, sports, and augmented reality.
Most previous works built their model via the 2D-3D lifting method~\cite{graformer, pavllo, stgcn}, which inferences the 2D pose in images by off-the-shelf 2D pose models ~\cite{cas} first, and then the 2D pose is taken as the input of lifting models to predict 3D pose. The method of separating 3D HPE into two phases can abate the influence of image backgrounds ~\cite{pavllo}.

3D HPE in the video has been developed for several years  ~\cite{pavllo, mhformer,mixste}. 
Temporal modeling regards temporal information as independent tokens, which demands performance computing to run these models ~\cite{mhformer,mixste, pstmo}.
Considering the computational cost, single-frame models can be easier for real-world applications. 

Early work has shown that features transformed from 2D poses can be useful information to estimate 3D positions ~\cite{pavllo}.
To solve the lack of capturing the spatial relationships among joints, graph convolutional networks (GCNs) have recently been adopted in many HPE models~\cite{stgcn, xu, liu, Zou}. 
 A drawback of the GCNs derived from spectrum convolution is weight-sharing convolution.
Each node in the GCNs is transformed by the same transformation matrix, and then the neighboring features will be aggregated to transfer information to the next layer ~\cite{kip}.
The weight-sharing method may not work well to capture the information on human joints because the flexibility and speed of human motion vary with joints \cite{stgcn,liu}.
Global dependency among joints in single-frame 3D HPE remains unclear. 
The Transformer-based models in computer vision have recently shown high performance in various tasks ~\cite{vit}. Since the self-attention mechanism in the Transformer can capture the global information among joints~\cite{vit}. In the case of 3D HPE, the self-attention mechanism can relate each joint to all the other joints to obtain global dependency~\cite{mixste}. The self-attention mechanism is suitable to model the similarity relations among joints since the global dependence can alter with the different input poses~\cite{graformer, poseformer}. In spite of that, the purely Transformer-based models may lack the physical information in the human skeleton ~\cite{graformer}. To address this issue, the previous works ~\cite{graformer, ijcai} proposed integrating multi-hop graph convolution into the Transformer. However, both the graph convolution with multi-hop range~\cite{chebnet} and self-attention have large receptive fields, which may not explicitly capture the local relations in the human skeleton. Thus we adopt the GCNs ~\cite{kip} to model the physically connected joints to improve the effectiveness of the model for 3D HPE. Additionally, following the success of the previous structures ~\cite{mixste, exploit, graphmlp}, we propose to alternately mix the local and global information by two independent modules, which are the GCN block and Transformer encoder, respectively.

Contributions in this paper can be summarized as follows: 1) On the basis of the global-local attention structure, we propose to alternately stack the transformer encoders and GCN blocks, namely AMPose, to capture the global and local information for 3D HPE. 2) Different designs are explored for the GCN blocks, which can fully exploit the physical-connectivity features. 3) Our model outperforms the state-of-the-art models on the Human3.6M dataset. The proposed model also shows better generalization ability on MPI-INF-3DHP as compared to previous works.



 


\vspace{-5mm}

\section{Methodology}
\label{sec:Methodology}
\vspace{-0.2cm}
\subsection{Overview}
\begin{figure}[t]
\centering 
\includegraphics[width=0.48\textwidth]{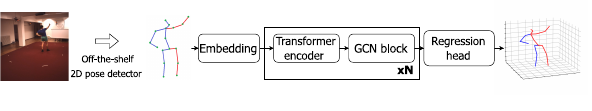} 
\caption{Overview of the AMPose network structure.}
\label{img1} 
\vspace{-0.2cm}
\end{figure}

The overall structure of the AMPose is illustrated in Fig. \ref {img1}. In this work, we aim to predict the 3D joint position. Our model takes 2D joint positions estimated by the off-the-shelf 2D pose model ~\cite{cas} as input first. 

Each joint pose will be transformed by a trainable linear projection layer, and then we alternately stack the Transformer encoder and the GCN block to transform the embedded features. The feature from the last GCN block is regressed to generate the 3D pose with a linear projection layer.

\subsection{Transformer encoder}
{\bf Scaled Dot-Product Attention} is defined as a self-attention function~\cite{vit}. 
This function takes query $Q$, key $K$ and value $V$ as input, where $Q, K, V$ $\in$ $R^{N_{j} \times N_{d}}$, $N_{j}$ is the number of joints and $N_{d}$ is the number of channels. 
To prevent the extremely big product of the query and the key, $\sqrt N_{d}$ is applied for normalization~\cite{vit}.
After the input feature $Z$ $\in$ $R^{N_{j} \times N_{d}}$ is transformed via learnable transformation matrix $W^{Q}$, $W^{K}$, $W^{V} \in R^{N_{d} \times N_{d}}$, the query, key, and value can be obtained, namely 
\begin{gather}
    Q = ZW^{Q}, K = ZW^{K}, V = ZW^{V}, 
\end{gather}
Self-attention mechanism can be formulated as follows:
\begin{gather}
    Attention\left(Q, K, V\right) = Softmax\left(\frac {QK^{T}} {\sqrt N_{d}}\right)V,
\end{gather}

\noindent {\bf Multi-head Self Attention Layer} (MSA) is the concatenation of multiple self-attention functions~\cite{mhformer}. 
The $Q$, $K$, and $V$ are split into $Q_{i}$, $K_{i}$, and $V_{i}$ $\in R^{N_{j} \times \frac {N_{d}} {N_{h}}}$, $i \in [1,...,N_{h}]$, respectively, 
where $N_{h}$ is the number of heads.
The concatenation of multiple heads will be transformed by $W^{out}$. The MSA can be represented as follows:
\begin{gather}
    MSA\left(Q, K, V\right) = concat\left( head_{1},...,head_{N_{h}}\right) W^{out},
\end{gather}
\noindent where $head_{i} =  Attention\left(Q_{i}, K_{i}, V_{i}\right)$ for all $i \in [1,...,N_{h}]$, and $W^{out} \in R^{N_{d} \times N_{d}} $ is the trainable parameter matrix.

\noindent {\bf Multi-layer perceptron (MLP)} is applied to project the feature of each joint with the fully connected layers (FCs). The Gelu is used as an activation function between FCs. The MLP can be expressed as follows:

\begin{gather}
    MLP(Z) = FC(\sigma_{Gelu} (FC(Z))),
\end{gather}
The overall process can be formulated as follows:
\begin{gather}
   Z' = Z_{0} + W_{pos}, \\
   Z'' = MSA(LN(Z')) + Z', \\
   Z_{1} = MLP(LN(Z'')) + Z'' ,
\end{gather}

\noindent where $Z_{0}$ denotes the input feature, $LN(\cdot)$ represents layer normalization, $W_{pos}$ is the positional embedding, and $Z_{1}$ denotes the output of the encoder.

\begin{table*}[ht] \small
  \centering
  \caption{MPJPE results of various actions in millimeters on Human3.6M. 
  Top table: 2D detector CPN with 17 keypoints used as input.  Bottom table: Ground truth 2D (GT) with 17 keypoints used as input.}
  \setlength{\tabcolsep}{0.8mm}{}
  \begin{tabular}{l c c c c c c c c c c c c c c c c}
    \hline
    CPN & Dire. & Discu. & Eat & Greet & Phone & Photo & Pose & Purchu. & Sit & SitD. & Smoke & Wait & WalkD. & Walk & WalkT. & Avg.\\ 
    \hline

Ci {\it et al.} ~\cite{ci} &  46.8 & 52.3 & {\bf 44.7} & 50.4 & 52.9 & 68.9 & 49.6 & 46.4 & 60.2 & 78.9 & 51.2 & 50.0 & 54.8 & 40.4 & 43.3 & 52.7 \\
Cai {\it et al.} (refinement) ~\cite{stgcn} &  46.5 & 48.8 & 47.6 & 50.9 & 52.9 & 61.3 & 48.3 & 45.8 & 59.2 & 64.4 & 51.2 & 48.4 & 53.5 & 39.2 & 41.2 & 50.6 \\ 
Pavllo {\it et al.} ~\cite{pavllo}  &  47.1 & 50.6 & 49.0 & 51.8 & 53.6 & 61.4 & 49.4 & 47.4 & 59.3 & 67.4 & 52.4 & 49.5 & 55.3 & 39.5 & 42.7 & 51.8 \\
Zou {\it et al.} (refinement) ~\cite{Zou} &  45.4 & 49.2 & 45.7 & 49.4 & {\bf 50.4} & 58.2 & 47.9 & 46.0 & 57.5 & {\bf 63.0} & 49.7 & 46.6 & 52.2 & 38.9 & 40.8 & 49.4\\
Lutz {\it et al.} ~\cite{jointformer} & 45.0 & 48.8 & 46.6 & 49.4 & 53.2 & 60.1 & 47.0 & 46.7 & 59.6 & 67.1 & 51.2 & 47.1 & 53.8 & 39.4 & 42.4 & 50.5 \\
\hline
Ours &  44.9 & 49.3 & 45.2 & 48.8 & 51.3 & 58.6 & 47.8 &  {\bf 44.8} & 57.1 & 66.5 & 49.9 & 46.4 & 52.9 & 39.0 & 40.6 & 49.5 \\
Ours (refinement) &  {\bf 42.8} & {\bf 48.6} & 45.1 & {\bf 48.0} & 51.0  &  {\bf 56.5}  & {\bf 46.2}  & 44.9  & {\bf 56.5}  & 63.9  & {\bf 49.6}  & {\bf 46.2}  & {\bf 50.5}  & {\bf 37.9}  & {\bf 39.5} & {\bf 48.5}
\\
\hline
  \end{tabular}
  
  \label{tab:example}

\begin{tabular}{l c c c c c c c c c c c c c c c c}

    \hline
    GT & Dire. & Discu. & Eat & Greet & Phone & Photo & Pose & Purchu. & Sit & SitD. & Smoke & Wait & WalkD. & Walk & WalkT. & Avg.\\ 
    \hline


Ci {\it et al.} ~\cite{ci} & 36.3 & 38.8 & 29.7 & 37.8 & 34.6 & 42.5 & 39.8 & 32.5 & 36.2 & {\bf 39.5} & 34.4 & 38.4 & 38.2 & 31.3 & 34.2 & 36.3  \\
Lutz {\it et al.} ~\cite{jointformer} & 31.0 & 36.6 & 30.2 & 33.4 & {\bf 33.5} &  39.0 & 37.1 & 31.3 & 37.1 & 40.1 & 33.8 & {\bf 33.5} & 35.0 & 28.7 & 29.1 & 34.0 \\
Zeng {\it et al.} ~\cite{zeng} & 32.9 & {\bf 34.5} & {\bf 27.6} & {\bf 31.7} & {\bf 33.5} & 42.5 & {\bf 35.1} & {\bf 29.5} & 38.9 & 45.9 & 33.3 & 34.9 & 34.4 & {\bf 26.5} & {\bf 27.1} & 33.9 \\
\hline

Ours & {\bf 30.7} & 35.9 & 30.0 & 33.2 & 33.9 & {\bf 37.7} & 37.1 & 31.9 & {\bf 34.3} &  41.8 & {\bf 33.2} & 34.5 & {\bf 34.3} &  27.5 & 28.5  & {\bf 33.7}
 \\
\hline
\end{tabular}
\vspace{-0.4cm}
  
\label{tab:example1}
\end{table*}
 
\begin{figure} [t]
\centering 
\includegraphics[width=0.48\textwidth]{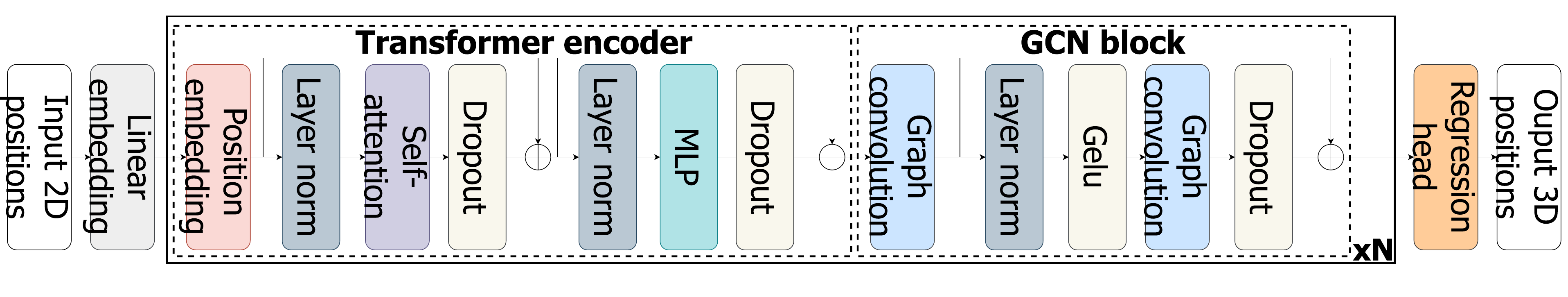}
\caption{An instantiation of the AMPose for 3D HPE.}
\label{img2} 
\vspace{-0.3cm} 
\end{figure}

\subsection{GCN blocks}

In these blocks, a spectral graph convolution ~\cite{chebnet, kip} is used to model the physical relations in the human skeleton since it can incorporate the information of structured data with a handcrafted graph~\cite{stgcn, xu}.
The Chebnet ~\cite{chebnet} derived from spectral graph convolution is a $k$-th order polynomial on an adjacency matrix. The high-order polynomial results in overfitting and higher computational complexity~\cite{kip}.
Thus Kipf {\it et al.} ~\cite{kip} simplify the spectral graph convolution with the first-order approximation, which also shortens the receptive field of the filter to the 1-hop range. The formula of the vanilla GCNs is as follows:
\begin{gather}
    Z = D^{-0.5}AD^{-0.5}X\Theta,
\end{gather}

\noindent where $A \in \{0,1\}^{N_{j} \times N_{j}}$ denote the adjacency matrix which includes the self-loop edges for all nodes,  
the diagonal matrix $D \in R^{N_{j} \times N_{j}}$ is the degree matrix of $A$, $X$ $\in R^{N_{j} \times N_{d}}$ is the input signal,
$\Theta$ $\in R^{N_{d} \times N_{D}}$ is the filter parameters, $Z \in R^{N_{j} \times N_{D}}$ is the output signal, $N_{j}$ is the number of nodes in the graph, $N_{d}$ is the input dimension of the signal, and $N_{D}$ is the output dimension of the signal.

The vanilla GCNs are originally proposed to address the node classification problem ~\cite{stgcn}. 
The fully weight-sharing operation can be used to solve the classification problem ~\cite{liu}, 
but the HPE comprehends complex human motion and different flexibility for each joint.
Prior works ~\cite{seman, stgcn} proposed to describe semantic relations among joints with more parameters to solve these problems. In our work, we adopt the similar design method as ST-GCN ~\cite{stgcn}, but we only classify the joint nodes into three groups based on the idea of physical and local relationships in the human skeleton: 
1) The center joint itself. 2) Neighboring joints which have a less number of hops to the mid hip than the center joint. 3) Neighboring joints which have a higher number of hops to the mid hip than the center joint. 

\noindent The formula of graph convolution is modified accordingly: The different
feature transformations are applied to different joints according to the three groups, and then the transformed feature will be aggregated, namely
\begin{gather}
    Z = \sum_{k=1}^{3} D_{k}^{-0.5}A_{k}D_{k}^{-0.5}X\Theta_{k},
\end{gather}
\noindent where $k$ denotes the index of different groups, the adjacent matrix $A \in \{0,1\}^{N_{j} \times N_{j}}$ is divided into three sub-matrices $A_{k} \in \{0,1\}^{N_{j} \times N_{j}}$ according to the joint groups which satisfy $A_{1}+ A_{2}+A_{3}=A$, the diagonal matrix $D_{k}\in R^{N_{j} \times N_{j}}$ is the degree matrix of $A_{k}$, and $\Theta_{k}$ is filter parameters for the $k$-th group. 
\subsection{Loss function}
To predict the 3D position with the AMPose, the mean square error is used as the loss function of our model:
\begin{gather}
    Loss = \frac 1 N \sum_{i=1}^{N} \left\Vert \mathbf{X_{i}} - \tilde{\mathbf{X_{i}}} \right\Vert^2,
\end{gather}
\noindent where $N$ is the number of the human joints, $i$ represents the index of joint type, $X_{i}$ denotes the 3D ground truth positions, and $\tilde{X_{i}}$ denotes the predicted 3D positions.

\section{Experiments}

\subsection{Dataset}
The AMPose is tested on two public datasets: Human3.6M ~\cite{human3.6} and MPI-INF-3DHP ~\cite{mpi}.

\noindent {\bf Human3.6M} is one among the most common datasets for 3D HPE,  
which is composed of 3.6 million images. The images are taken from 4 cameras arranged at different angles.
Following previous work ~\cite{Zou,graformer}, the 2D joint positions are normalized as the input of the AMPose. The mid hip is adopted as the root joint of the 3D pose, which localizes the 3D human positions. The AMPose is trained with subjects S1, S5, S6, S7, and S8, while tested on subjects S9 and S11.

\noindent {\bf  MPI-INF-3DHP} is a challenging dataset due to its various conditions. The dataset is composed of images with indoor and outdoor backgrounds along with complex and rare actions. Following prior work ~\cite{xu, Zou}, we evaluate the generalization ability of the AMPose by training on Human3.6M and testing on the test set of MPI-INF-3DHP. In addition, we also trained the AMPose with the training set of MPI-INF-3DHP and tested it on the test set of MPI-INF-3DHP to compare with the video-based methods~\cite{pstmo, mixste, mhformer}.


\begin{figure}[ht!] 
\centering 
\includegraphics[width=0.35\textwidth]{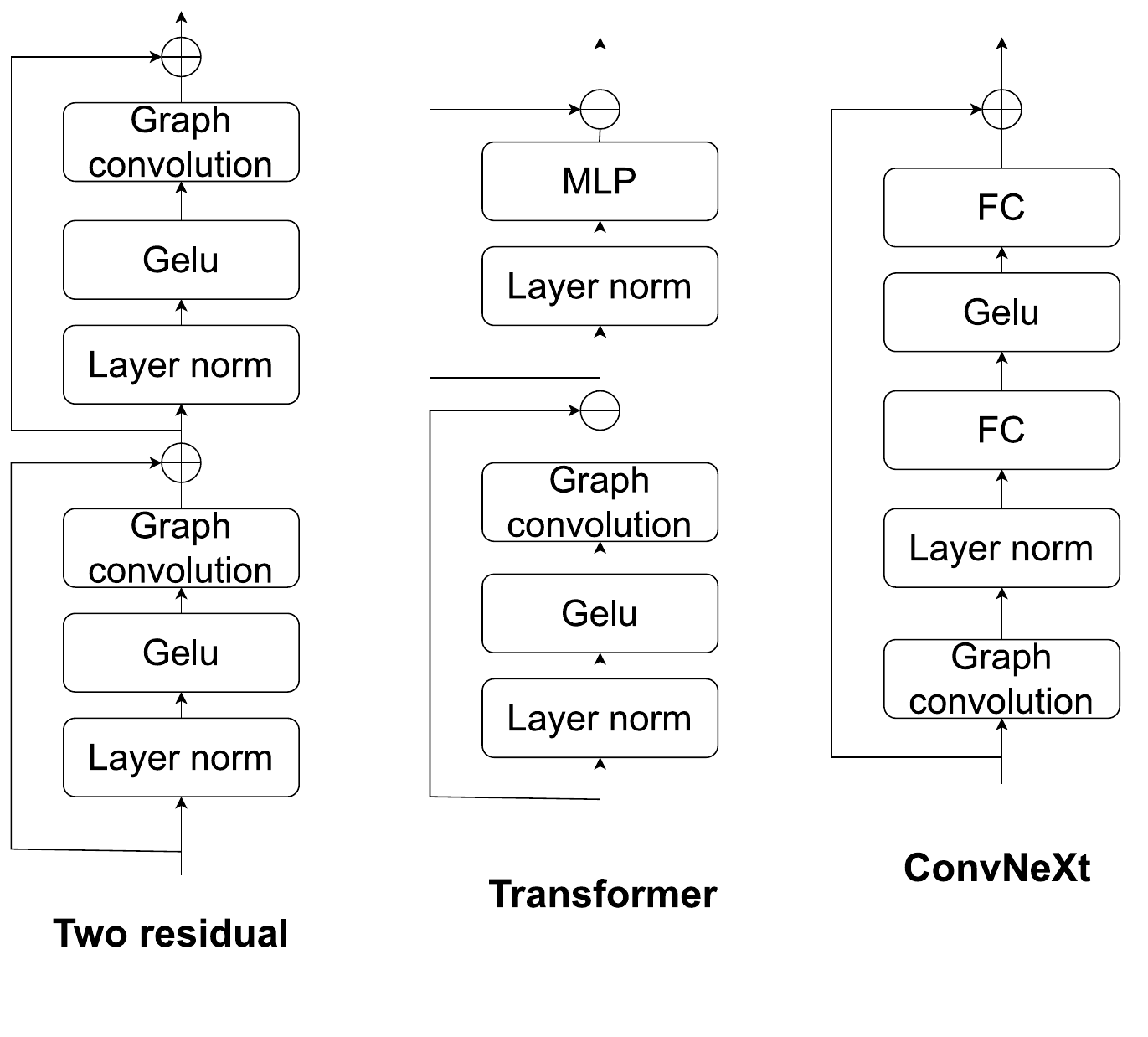}
\caption{Various designs for the GCN block.} 
\label{img3} 
\vspace{-0.4cm}
\end{figure}

\subsection{Evaluation protocol}
The mean per joint position error (MPJPE) is used as the evaluation protocol. 
MPJPE is computed as the mean of Euclidean distance between the estimated 3D positions and the ground truth positions.

For cross-dataset comparison, two metrics are used to gauge the generalization ability of the proposed model: the Percentage of Correct Keypoints (PCK) and the Area Under the Curve (AUC). 
The estimated 3D pose is regarded as the correct pose if MPJPE is below the predefined threshold.
Following prior works ~\cite{Zou,liu, xu}, the threshold is set as 150 millimeters (mm).

\begin{table}[t] \small
\centering
\caption{MPJPE results in millimeters on the Human3.6M dataset. CPN and GT with 16 keypoints used as input.}
\setlength{\tabcolsep}{1mm}{}
\begin{tabular}{l c c c c c c c c c c c c c c c c}
\hline
Method  & Param. (M) &MPJPE (CPN) & MPJPE (GT) \\
\hline
Liu {\it et al.} ~\cite{liu}  & 4.22 & 52.4 & 37.8 \\
Xu {\it et al.} \cite{xu}  & 3.70 & 51.9 & 35.8 \\
Zhao {\it et al.} \cite{graformer}  & {\bf 0.65} & 51.8 & 35.2 \\
\hline
Ours (small)  & 0.67   & 51.3 & 34.8 \\
Ours (full)  & 18.3  & {\bf 50.5} & {\bf 34.1} \\
\hline
\end{tabular}
\label{tab:example7}
\end{table}

\begin{table}[ht] \small
\centering
\caption{Cross-dataset comparison on MPI-INF-3DHP.
}
\setlength{\tabcolsep}{1mm}{}

\begin{tabular}{l c c c c c c c c c c c c c c c c}

\hline
Method & GS & noGS &  Outdoor & Avg. PCK & AUC  \\
\hline
Ci {\it et al.} ~\cite{ci} &  74.8 & 70.8 & 77.3 & 74.0 & 36.7 \\
Zhao {\it et al.} ~\cite{graformer} & 80.1 & 77.9 & 74.1 & 79.0 & 43.8 \\
Liu {\it et al.} ~\cite{liu} & 77.6 & 80.5 & 80.1 & 79.3 & 47.6 \\
Xu {\it et al.} ~\cite{xu} & 81.5 & 81.7 & 75.2 & 80.1 & 45.8 \\
Zou {\it et al.} ~\cite{Zou} & {\bf 86.4} & 86.0 & 85.7 & 86.1 & 53.7 \\
\hline
Ours & 86.1 & {\bf 87.5} & {\bf 87.4} & {\bf 87.0}  & {\bf 55.2}\\
\hline
\end{tabular}
\label{tab:example3}
\end{table}

\begin{table}[t] \footnotesize
\centering
\caption{Comparison results on the MPI-INF-3DHP dataset. }

\setlength{\tabcolsep}{0.7mm}{}
\begin{tabular}{l c c c c c c c c c c c c c c c c}
\hline
Method &	Param. (M) &	FLOPs (M) &	PCK &	AUC &	MPJPE \\
\hline
Li {\it et al.} (9 frames) \cite{mhformer} &	18.9 &	1030 &	93.8  &	63.3  &	58.0 \\
Zhang {\it et al.} (1 frame) \cite{mixste} &	33.7 &	645 &	94.2 &	63.8  &	57.9  \\
Zhang {\it et al.} (27 frames) \cite{mixste} &	33.7 &	645	 & 94.4  &	66.5  &	54.9  \\
Shan {\it et al.} (81 frames) \cite{pstmo} &	{\bf 5.4}	 & 493 &	97.9 &	75.8	 & 32.2 \\
\hline
Ours (1 frame) &	18.3 &	{\bf 312}	 & {\bf 98.0} &	{\bf 79.1} &	{\bf 30.8} \\
\hline
\end{tabular}
\label{tab:example2}
\end{table}

\subsection{Implementation details}
The AMPose is built on the two-phase method. 
The images are first fed into the off-the-shelf 2D pose model, which is the cascaded pyramid network (CPN) ~\cite{cas} for our experiments on Human3.6M.

In our experiments, the depth $N$ of the proposed model is set to 5. The number of embedding channels is set to 512. 
The model is trained for 50 epochs. The batch size is set to 128. The learning rate is initially set to 0.000025 along with the Adam optimizer. 
All experiments are run on a machine equipped with a single NVIDIA RTX 2080 GPU.

\subsection{Comparison with the state-of-the-art}
Table 1 and Table 2 compare the AMPose with state-of-the-art single-frame models over CPN and ground truth keypoints on the Human3.6M dataset. Though some models additionally apply the pose refinement module ~\cite{stgcn} to improve the accuracy, we report the result before the refinement for fair comparisons. As shown in Table 1 and Table 2, the AMPose outperforms all prior methods. 
Table 3 compares the generalization ability by testing the model trained on Human3.6M with MPI-INF-3DHP. We separately report the result with different backgrounds, including green screen (GS), noGS, and outdoor. Our model reaches an 87\% PCK without fine-tuning on MPI-INF-3DHP. Additionally, we trained the AMPose on the training set of MPI-INF-3DHP to compare the video-based methods. As presented in Table 4, the proposed method outperforms the state-of-the-art methods with lower FLOPs.

\begin{table}[ht!]  \small
\centering
\caption{The result of the different designs for the GCN block.
}
\setlength{\tabcolsep}{0.8mm}{}

\begin{tabular}[h] { l c c c c }
\hline
Method & Param. (M) & FLOPs (M) & MPJPE (mm) & $\Delta$ \\
\hline
Two residual & 18.5 & 312.2  & 50.1 & 0.6\\
Transformer & 17.0 & 289.8 & 50.1 & 0.6\\
ConvNeXt & 17.0 & 289.9 & 50.0 & 0.5\\
\hline
Ours  & 18.3 & 312.2 & 49.5 & -\\
\hline
\vspace{-2em}
\end{tabular}

\label{tab:example9}
\end{table}

\subsection{Ablation Study}
We explore the optimal design of the GCN block for the proposed method. As depicted in Fig. \ref {img3}, we test the different designs of the GCN block which incorporate graph convolution into the Transformer ~\cite{vit}
and ConvNeXt~\cite{convnext}, respectively. The result and floating point operations (FLOPs) are shown in Table 5. Our design is shown in Fig. \ref {img2}. We found that a plain graph convolution followed by a graph convolution with the residual has the best result in our experiment.

\begin{figure}[ht!]
\center
\includegraphics[width=0.45\textwidth]{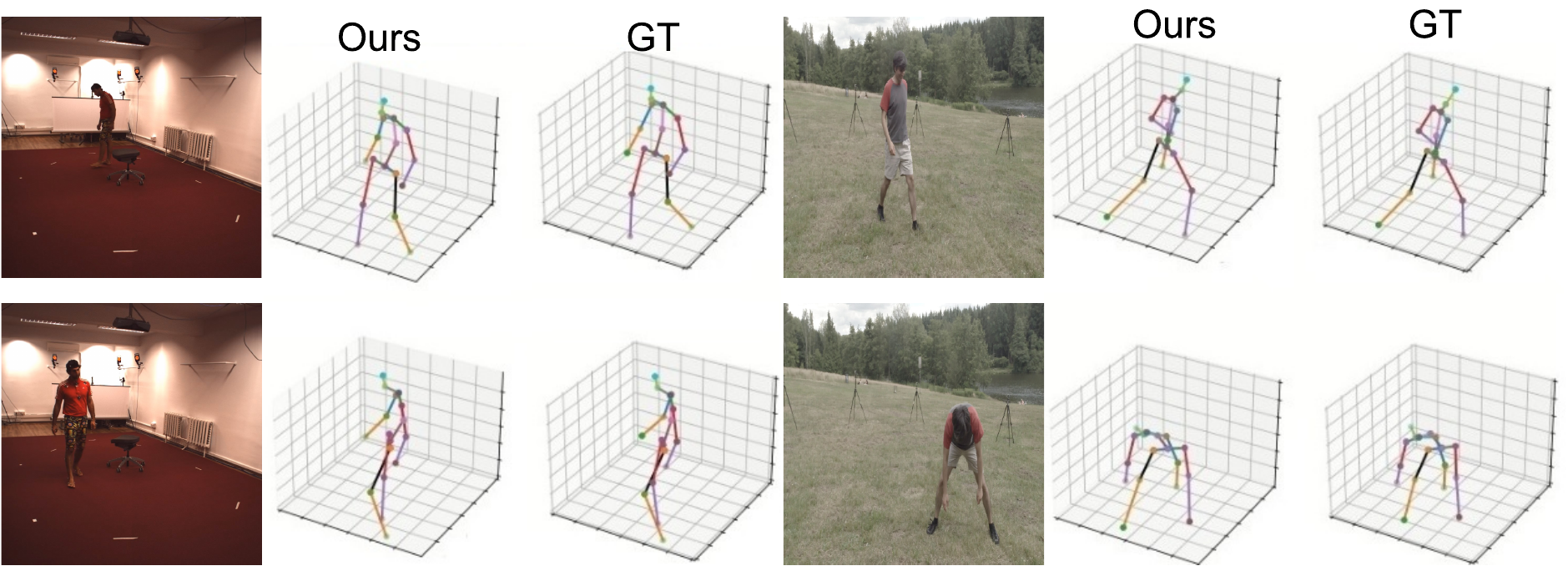}


\caption{Visualization on the test set of Human3.6M and MPI-INF-3DHP.}
\label{img4}
\end{figure}

\subsection{Qualitative Results}
Fig. \ref {img4} shows some visualization results estimated by the proposed model on Human3.6M and MPI-INF-3DHP. The AMPose correctly predicts 3D pose actors in various backgrounds. Even the actors performing rare actions in the wild can be accurately estimated by the proposed method. 


\section{Conclusion}
In this work, we have presented a novel method AMPose for 3D HPE in which the Transformer encoder and GCN blocks are alternately stacked. 
In the AMPose, the relations among joints are divided into two sets: The global and physically connected relations in human joints, which the Transformer encoder and GCNs can modularize, respectively. 
The proposed method shows superior performance in accuracy and generalization ability by comparing with state-of-the-art models on the Human3.6M and MPI-INF-3DHP datasets. 
For realistic applications, the lightweight HPE model based on the proposed structure may be future work.


\bibliographystyle{IEEEbib}


\begin{thebibliography}{10}

\bibitem{graformer}
Weixi Zhao, Weiqiang Wang, and Yunjie Tian,
\newblock ``Graformer: Graph-oriented transformer for 3d pose estimation,''
\newblock in {\em CVPR}, 2022, pp. 20406--20415.

\bibitem{pavllo}
Dario Pavllo, Christoph Feichtenhofer, David Grangier, and Michael Auli,
\newblock ``3d human pose estimation in video with temporal convolutions and
  semi-supervised training,''
\newblock in {\em CVPR}, 2019, pp. 7753--7762.

\bibitem{stgcn}
Yujun Cai, Liuhao Ge, Jun Liu, Jianfei Cai, Tat-Jen Cham, Junsong Yuan, and
  Nadia~Magnenat Thalmann,
\newblock ``Exploiting spatial-temporal relationships for 3d pose estimation
  via graph convolutional networks,''
\newblock in {\em ICCV}, 2019, pp. 2272--2281.

\bibitem{cas}
Yilun Chen, Zhicheng Wang, Yuxiang Peng, Zhiqiang Zhang, Gang Yu, and Jian Sun,
\newblock ``Cascaded pyramid network for multi-person pose estimation,''
\newblock in {\em CVPR}, 2018.

\bibitem{mhformer}
Wenhao Li, Hong Liu, Hao Tang, Pichao Wang, and Luc Van~Gool,
\newblock ``Mhformer: Multi-hypothesis transformer for 3d human pose
  estimation,''
\newblock in {\em CVPR}, 2022, pp. 13137--13146.

\bibitem{mixste}
Jinlu Zhang, Zhigang Tu, Jianyu Yang, Yujin Chen, and Junsong Yuan,
\newblock ``Mixste: Seq2seq mixed spatio-temporal encoder for 3d human pose
  estimation in video,''
\newblock in {\em CVPR}, 2022, pp. 13222--13232.

\bibitem{pstmo}
Wenkang Shan, Zhenhua Liu, Xinfeng Zhang, Shanshe Wang, Siwei Ma, and Wen Gao,
\newblock ``P-stmo: Pre-trained spatial temporal many-to-one model for 3d human
  pose estimation,''
\newblock in {\em ECCV}, 2022, pp. 461--478.

\bibitem{xu}
Tianhan Xu and Wataru Takano,
\newblock ``Graph stacked hourglass networks for 3d human pose estimation,''
\newblock in {\em CVPR}, 2021, pp. 16100--16109.

\bibitem{liu}
Kenkun Liu, Rongqi Ding, Zhiming Zou, Le~Wang, and Wei Tang,
\newblock ``A comprehensive study of weight sharing in graph networks for 3d
  human pose estimation,''
\newblock in {\em ECCV}, 2020, pp. 318--334.

\bibitem{Zou}
Zhiming Zou and Wei Tang,
\newblock ``Modulated graph convolutional network for 3d human pose
  estimation,''
\newblock in {\em ICCV}, 2021, pp. 11457--11467.

\bibitem{kip}
Thomas~N Kipf and Max Welling,
\newblock ``Semi-supervised classification with graph convolutional networks,''
\newblock {\em arXiv preprint arXiv:1609.02907}, 2016.

\bibitem{vit}
Alexey Dosovitskiy, Lucas Beyer, Alexander Kolesnikov, Dirk Weissenborn,
  Xiaohua Zhai, Thomas Unterthiner, Mostafa Dehghani, Matthias Minderer, Georg
  Heigold, Sylvain Gelly, et~al.,
\newblock ``An image is worth 16x16 words: Transformers for image recognition
  at scale,''
\newblock {\em arXiv preprint arXiv:2010.11929}, 2020.

\bibitem{poseformer}
Ce~Zheng, Sijie Zhu, Matias Mendieta, Taojiannan Yang, Chen Chen, and Zhengming
  Ding,
\newblock ``3d human pose estimation with spatial and temporal transformers,''
\newblock in {\em ICCV}, 2021, pp. 11636--11645.

\bibitem{ijcai}
Yiran Zhu, Xing Xu, Fumin Shen, Yanli Ji, Lianli Gao, and Heng~Tao Shen,
\newblock ``Posegtac: Graph transformer encoder-decoder with atrous convolution
  for 3d human pose estimation,''
\newblock in {\em IJCAI}, 2021, pp. 1359--1365.

\bibitem{chebnet}
Micha{\"e}l Defferrard, Xavier Bresson, and Pierre Vandergheynst,
\newblock ``Convolutional neural networks on graphs with fast localized
  spectral filtering,''
\newblock {\em Advances in neural information processing systems}, vol. 29,
  2016.

\bibitem{exploit}
Wenhao Li, Hong Liu, Runwei Ding, Mengyuan Liu, Pichao Wang, and Wenming Yang,
\newblock ``Exploiting temporal contexts with strided transformer for 3d human
  pose estimation,''
\newblock {\em IEEE Transactions on Multimedia}, 2022.

\bibitem{graphmlp}
Wenhao Li, Hong Liu, Tianyu Guo, Hao Tang, and Runwei Ding,
\newblock ``Graphmlp: A graph mlp-like architecture for 3d human pose
  estimation,''
\newblock {\em arXiv preprint arXiv:2206.06420}, 2022.

\bibitem{ci}
Hai Ci, Chunyu Wang, Xiaoxuan Ma, and Yizhou Wang,
\newblock ``Optimizing network structure for 3d human pose estimation,''
\newblock in {\em ICCV}, 2019, pp. 2262--2271.

\bibitem{jointformer}
Sebastian Lutz, Richard Blythman, Koustav Ghosal, Matthew Moynihan, Ciaran
  Simms, and Aljosa Smolic,
\newblock ``Jointformer: Single-frame lifting transformer with error prediction
  and refinement for 3d human pose estimation,''
\newblock {\em arXiv preprint arXiv:2208.03704}, 2022.

\bibitem{zeng}
Ailing Zeng, Xiao Sun, Fuyang Huang, Minhao Liu, Qiang Xu, and Stephen Lin,
\newblock ``Srnet: Improving generalization in 3d human pose estimation with a
  split-and-recombine approach,''
\newblock in {\em ECCV}, 2020, pp. 507--523.

\bibitem{seman}
Long Zhao, Xi~Peng, Yu~Tian, Mubbasir Kapadia, and Dimitris~N Metaxas,
\newblock ``Semantic graph convolutional networks for 3d human pose
  regression,''
\newblock in {\em CVPR}, 2019, pp. 3420--3430.

\bibitem{human3.6}
Catalin Ionescu, Dragos Papava, Vlad Olaru, and Cristian Sminchisescu,
\newblock ``Human3. 6m: Large scale datasets and predictive methods for 3d
  human sensing in natural environments,''
\newblock {\em IEEE TPAMI}, vol. 36, no. 7, pp. 1325--1339, 2014.

\bibitem{mpi}
Dushyant Mehta, Helge Rhodin, Dan Casas, Pascal Fua, Oleksandr Sotnychenko,
  Weipeng Xu, and Christian Theobalt,
\newblock ``Monocular 3d human pose estimation in the wild using improved cnn
  supervision,''
\newblock in {\em 3DV}, 2017.

\bibitem{convnext}
Zhuang Liu, Hanzi Mao, Chao-Yuan Wu, Christoph Feichtenhofer, Trevor Darrell,
  and Saining Xie,
\newblock ``A convnet for the 2020s,''
\newblock in {\em CVPR}, 2022, pp. 11966--11976.

\end{thebibliography}

\end{document}